%% file: main.tex
\documentclass[10pt,twocolumn,letterpaper]{article}

\usepackage[pagenumbers]{cvpr}     
\input{preamble}
\definecolor{cvprblue}{rgb}{0.21,0.49,0.74}
\usepackage[pagebackref,breaklinks,colorlinks,allcolors=cvprblue]{hyperref}

\def\paperID{*****} 
\def\confName{CVPR}
\def\confYear{2026}

\title{Counting Animals in Camera-Traps Image Sequences without Count Labels:
Winning Solution to the iWildCam 2021 Challenge}

\author{Fagner Cunha, Juan G. Colonna, Eulanda M. dos Santos\\
Federal University of Amazonas\\
Manaus, Amazonas, Brazil\\
{\tt\small \{fagner.cunha, juancolonna, emsantos\}@icomp.ufam.edu.br}
}

\begin{document}
\maketitle
\input{sec/0_abstract}    
\input{sec/1_intro}
\input{sec/2_methods}
\input{sec/3_results}
\input{sec/4_conclusion}
{
    \small
    \bibliographystyle{ieeenat_fullname}
    \bibliography{main}
}


\end{document}

%% file: sec/0_abstract.tex
\begin{abstract}
Camera traps have become an essential tool for wildlife monitoring, motivating
the development of computer vision methods for the automated extraction of
information from these data. While most prior work has focused on species
identification, many ecological applications also require estimating the number
of unique individuals appearing across short image sequences. This task is
particularly challenging because camera traps typically acquire bursts of
images at approximately one frame per second, creating large temporal
discontinuities that may make conventional multi-object tracking methods
unreliable, and because manually collecting individual count annotations is
prohibitively expensive. In this work, we describe the winning solution to the
iWildCam 2021 Challenge, which introduced a benchmark for counting animals at
the sequence level under realistic annotation constraints where count
annotations are unavailable for training. Our approach, \textit{MaxBoxCount},
combines a strong species classification pipeline with a simple yet effective
counting heuristic based on MegaDetector detections to estimate the number of
unique individuals without requiring count annotations.  Code is available at
\url{https://github.com/alcunha/iwildcam2021ufam}.
\end{abstract}

%% file: sec/1_intro.tex
\section{Introduction}
\label{sec:intro}

Camera traps have been widely used for wildlife monitoring, collecting large
amounts of images worldwide \citep{ahumada2013monitoring, he2016visual}. Over
the last decade, the computer vision community has investigated a wide range of
approaches to improve the automated extraction of information from these data,
with a primary focus on species identification
\citep{norouzzadeh2018automatically, tabak2019machine,
willi2019identifying, schneider2020three, beery2023wild}. However, solely
identifying the species is not sufficient certain ecological modeling tasks,
such as estimating the abundance or density of the species
\citep{rowcliffe2013clarifying}. In these cases, counting the number of
individuals captured across image sequences is also required. Motivated by this
problem, the iWildCam 2021 Challenge \citep{beery2021iwildcam} introduced a
benchmark for counting animals across short camera-trap image sequences,
encouraging the development of methods that can estimate sequence-level counts
under realistic annotation constraints.

Unlike traditional object counting, the competition required estimating the
number of unique individuals from each species appearing across an entire
sequence rather than counting detections independently in each image. Camera
traps typically acquire bursts of images at approximately one frame per second
\citep{he2016visual, beery2018recognition}, creating large temporal
discontinuities that may make conventional multi-object tracking methods
unreliable. Furthermore, no count annotations were provided for the training
set, reflecting a realistic scenario in which manually collecting individual
counts is prohibitively expensive. Instead, participants received species labels
together with weakly supervised object detections and instance segmentations,
requiring solutions capable of estimating counts without direct supervision.

This formulation encouraged approaches that combine multiple sources of
information rather than relying on supervised counting models. A naïve strategy
based solely on the number of detections tends to overestimate counts when the
same individual appears in multiple images and underestimate them when different
individuals appear in different frames. Accurately estimating sequence-level
counts therefore requires reasoning about the correspondence of detections
across the image burst despite sparse temporal information.

In this work, we describe the winning solution to the iWildCam 2021 counting
challenge. Rather than proposing a learned counting model, our method combines
a strong species classification pipeline with a simple yet effective
sequence-level counting heuristic based on MegaDetector detections to
estimate the number of unique individuals without requiring count annotations.
Although designed specifically for the competition setting, the proposed
approach illustrates how existing computer vision components can be effectively
combined to solve weakly supervised counting problems in camera-trap image
sequences.

%% file: sec/2_methods.tex
\begin{figure*}[t]
    \centering
    \includegraphics[width=0.85\textwidth]{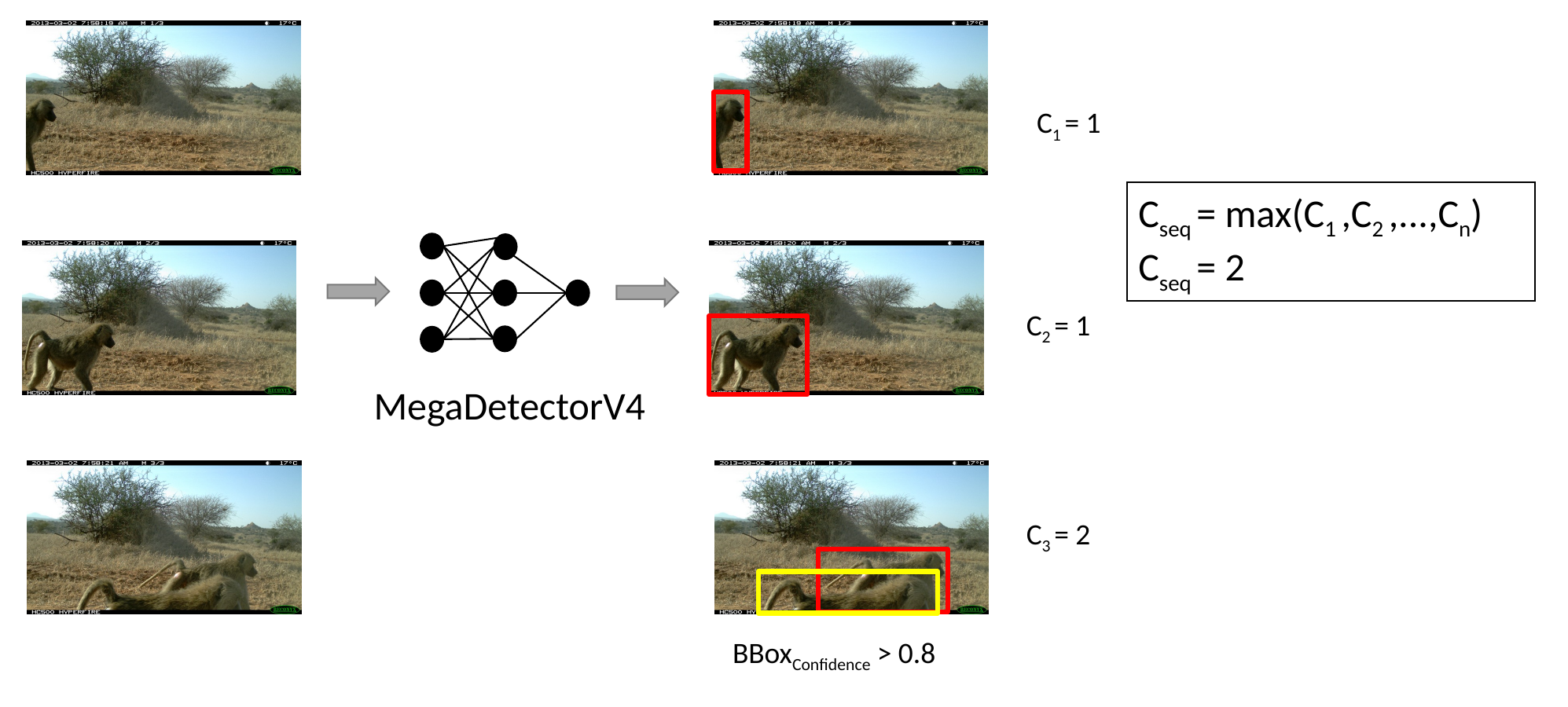}
    \caption{The MaxBoxCount heuristic estimates the count for a sequence as the
    number of bounding boxes in the image with the highest number of
    detections. Only bounding boxes with confidence greater than $\tau$
    are considered. For our iWildCam 2021 Challenge submission, we set
    $\tau = 0.8$.}
    \label{fig:baseline_count}
\end{figure*}

\section{Materials and Methods}
\label{sec:methods}

In this section, we present the dataset proposed by the iWildCam 2021 challenge
and describe our heuristic strategy, which serves as a strong baseline for the
task of counting animals in camera-trap images.

\subsection{The {iWildcam} 2021 dataset}

The iWildCam 2021 dataset comprises three components: camera trap
images from the Wildlife Conservation Society (WCS) dataset, citizen science images from iNaturalist, and
multispectral imagery for the camera trap locations. The camera trap data,
provided by the WCS~\citep{wcs_lila}, contains
263,528 images of 206 species from 414 locations across 12 countries worldwide.
The official training/test split is based on camera location, with the training
set containing 203,314 images from 323 locations and the test set containing
60,214 images from 91 locations. As most publicly available camera trap data
does not include count labels, the count labels for image bursts were collected
exclusively for the test set to encourage the development of methods that can
learn to count without explicit labels \citep{beery2021iwildcam}.

The dataset also includes a subset of the iNaturalist 2017-2019 competition
dataset \citep{horn2018inat}, with 13,051 additional images from 75 species.
Although these images typically present a different data distribution and are of
higher quality, they can still be valuable for improving classification.

Finally, the dataset includes raw remote sensing data for each camera location,
collected by the Landsat 8 satellite, allowing investigation into whether the
model performance can be improved using this kind of multimodal data. GPS
coordinates for each camera location are provided, with most of them obfuscated
for privacy and security reasons. Each location has been randomly adjusted to
lie within 1 km of the original location center.

\subsection{MaxBoxCount: A heuristic for counting animals in camera-trap
images}

Our proposed heuristic consists of two components: an image burst classifier
based on an ensemble of EfficientNet-B2 \citep{tan2019efficientnet} models and a
counting heuristic based on the bounding boxes generated by MegaDetectorV4
\citep{beery2019efficient}. The main idea is to estimate the number of animals
present in the sequence based on the image with the highest number of bounding
boxes (MaxBoxCount), which could provide a lower bound on the number of animals
across the sequence if the animal detector achieves perfect detection (see
\autoref{fig:baseline_count}). Additionally, since the vast majority of the
images contain only a single species, the simplest approach for classification
is to assume a single species across the entire sequence and use the most likely
species prediction for the total count of individuals.

\begin{figure*}[t]
    \centering
    \includegraphics[width=0.85\textwidth]{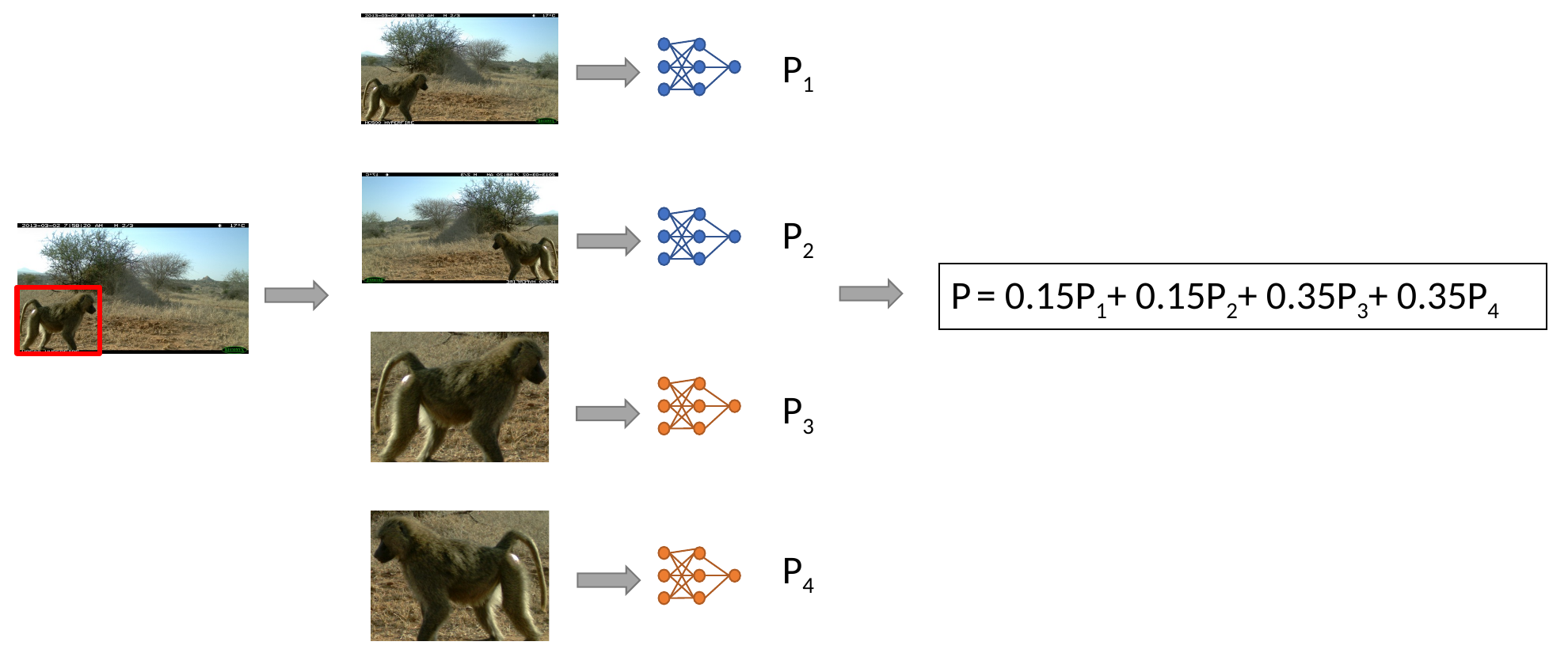}
    \caption{Fusion strategy for the classifier ensemble used in
    MaxBoxCount. The species prediction $P$ for a given image is computed as
    the weighted average of the predictions for the full image ($P_1$), its
    horizontally flipped version ($P_2$), the highest-confidence bounding box
    ($P_3$), and its horizontally flipped version ($P_4$).}
    \label{fig:baseline_class_ens}
\end{figure*}

Initially, MegaDetectorV4 is run on all images to detect animals. Next, we run
the image classifiers on both the full image and the bounding box with the
highest confidence score. The image prediction is calculated as the weighted
average of the predictions from both models, following the approach used by the
winning solution of iWildCam 2020: 0.15 (full image) + 0.15 (mirrored full
image) + 0.35 (bbox) + 0.35 (mirrored bbox). \autoref{fig:baseline_class_ens}
illustrates this fusion strategy. To predict the species in the sequence,
we average the species predictions of the non-empty images in the burst.

\begin{figure*}[t]
    \centering
\includegraphics[width=0.9\textwidth]{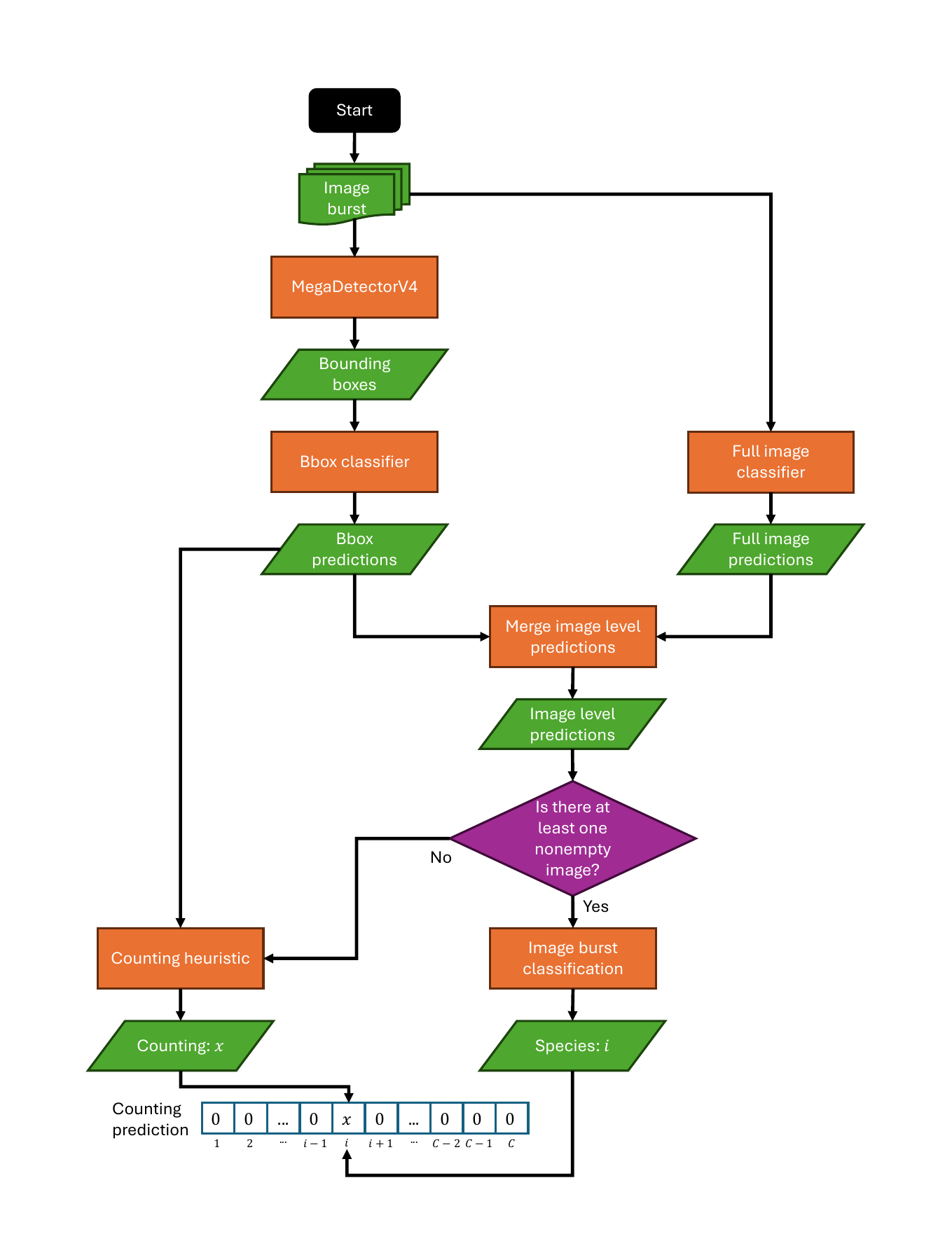}
    \caption{Overview of the MaxBoxCount heuristic baseline for animal
    counting. The method uses bounding boxes generated by MegaDetectorV4 to
    estimate the number of individuals in each image sequence. Species
    identification is performed using two classifiers: one operating on the
    full image and another on the detected bounding boxes.}
    \label{fig:wildcam2021_winning}
\end{figure*}

Finally, we generate a prediction vector containing the number of individuals of
each species -- in our case, only a single species -- based on the maximum
number of bounding boxes with confidence score higher than a threshold $\tau$
across any image in the sequence. If the classifiers identify only empty images,
the vector will contain only zeros. \autoref{fig:wildcam2021_winning}
presents a diagram summarizing the steps of our proposed heuristic baseline.

This classification strategy limits us to predicting only one species per
sequence. However, it has proven to be a strong baseline, as it was the winning
solution in iWildCam 2021. In the next subsection, we detail our
implementation.

\begin{table*}[t]
    \centering
    \caption{Training hyperparameters for each stage of the classifier training
process.}
    \label{tab:counting_baseline_train_stages}
    \begin{tabular}{l l c c l c}
        \toprule
        \textbf{Stage} & \textbf{Weights Trained} & \textbf{Epochs} &
        \textbf{LR} & \textbf{Preprocessing} & \textbf{Resolution} \\
        \midrule
        Stage 1$^{*}$ & Classifier layer         & 4  & 0.01  & Training  &
$260\times260$ \\
        Stage 2       & All                      & 20 & 0.01  & Training  &
$260\times260$ \\
        Stage 3       & Last block + classifier  & 2  & 0.001 & Inference &
$380\times380$ \\
        Stage 4       & BAGS head                & 12 & 0.01  & Training  &
$380\times380$ \\
        Stage 5       & BAGS head                & 2  & 0.001 & Inference &
$380\times380$ \\
        \bottomrule
    \end{tabular}

    \vspace{2pt}
    \footnotesize{$^{*}$ Applied only to the bounding-box classifier.}
\end{table*}

\subsection{Species classifier}\label{sec:iwildcam_class}

We trained two species classifiers based on the EfficientNet-B2 architecture:
one trained using full images and another (bbox) trained with square crops
containing animals, extracted from bounding boxes generated by MegaDetectorV4.
We chose this approach following the iWildCam 2020 solutions, where the bbox
model can provide better predictions due to the highlighted animal. On the other
hand, the full image model has been shown to be more effective at classifying
images with animal herds. We used only the provided camera-trap data to train
the classifier models.

To train the bbox model, we treated all bounding boxes with a confidence score
equal to or higher than 0.6\footnote{The threshold of 0.6 for including a bounding box in the training set 
was selected based on a held-out validation set. Note that this detection 
confidence threshold is distinct from the counting threshold $\tau = 0.8$ 
used during inference.} as independent training samples and applied square crops
around them, using the size of the largest bbox side. If no bounding box met
this confidence threshold, we used the full image for that training instance.
The image label was assigned as the bounding box label.

During inference, the species prediction assigned to each image is the weighted
average of the predictions from the bounding box with the highest score, the
full image, and their horizontally mirrored versions, following the approach
used by the winning solution of iWildCam 2020: 0.15 (full image) + 0.15
(mirrored full image) + 0.35 (bbox) + 0.35 (mirrored bbox). To classify a
sequence, we average the predictions of all nonempty images within the sequence.

\textbf{Handling class imbalance}. To address the class imbalance problem
inherent in real-world applications, such as the one represented in iWildCam
2021 dataset, we applied the Balanced Group Softmax (BAGS) method
\citep{li2020overcoming}. Following the original BAGS paper, we grouped classes
into 4 softmax groups based on the number $N$ of training instances: $N < 10$,
$10 \le N < 100$, $100 \le N < 1000$, $N \ge 1000$. For each softmax group, we
included the class ``others'' to represent instances from all classes not
included in that group. During training, for each softmax, we undersampled the
``others'' category to ensure it contained at most eight times the number of
instances belonging to other classes per batch, to prevent it from dominating
the softmax. Additionally, we included a special softmax group for the
foreground/background classification. For the final prediction, we remapped all
predictions to the original softmax, ignoring the ``others'' class. As in the
original paper, these predictions are not true probabilities since they do not
sum up to one, but we consider the highest value as the BAGS prediction.

\textbf{Image preprocessing}. For image preprocessing, we applied a square crop
centered on the bounding boxes or a random crop of aspect ratio
sampled in $[3/4, 4/3]$ and area in $[65\%, 100\%]$ for full image, followed
by a random horizontal flip and RandAugment (N=6, M=2). Then, the images were
normalized and resized to match the input dimensions of the network. During the
fix train/test resolution stage, we used inference preprocessing, which consists
only of normalizing and resizing the image or square crop to the input
dimensions of the network.

\textbf{Multi-stage training}. We trained the models in multiple stages. First,
each model was trained using the standard softmax with the default input
resolution of EfficientNet-B2 ($260 \times 260$). In the first stage, only the
classifier layer was trained for 4 epochs, while the backbone, previously
initialized with ImageNet weights, remained frozen. In the second stage, all
layers were unfrozen, and all weights were fine-tuned for 20 epochs. In the
third stage, to fix the train/test resolution discrepancy issue
\citep{touvron2019fixing}, we fine-tuned only the last inverted bottleneck block
of the EfficientNet-B2 architecture and the classifier layer for 2 more epochs.
This was done using a higher input resolution ($380 \times 380$) and applying the
inference preprocessing steps to the images. Afterward, we removed the plain
softmax and trained the BAGS head on top of the backbone, which was kept
frozen. In the fourth stage, the BAGS header was trained for 12 epochs using
training preprocessing with data augmentation at the inference resolution ($380 \times 380$). Finally, in the fifth stage, the BAGS header was fine-tuned for 2 epochs
using inference-time preprocessing. \autoref{tab:counting_baseline_train_stages}
presents a summary of the parameters used.

\textbf{Implementation details}. To adjust the training hyperparameters, we
held out a validation set from the training set based on the locations.
However, to train the final models, we used all images available in the
original training set. The models were initialized with ImageNet weights and
trained with a batch size of $32$ using SGD with an initial learning rate of
$0.01$ ($0.001$ during the fix train/test resolution stages) and momentum of
$0.9$. The learning rate was linearly warmed up from $0$ to the initial value
over one-third of the steps in the first epoch of each training stage, and then
decayed to $0$ following the cosine schedule. We also applied label smoothing
with a value of $0.1$.

%% file: sec/3_results.tex
\section{Results}\label{sec:iwildcam_results}

To evaluate animal counting methods, the iWildCam 2021 Challenge uses the
mean column-wise root-mean-squared error (MCRMSE). Let
$\mathbf{Y} \in \{0,1,2,\dots\}^{n \times m}$ denote the matrix of ground-truth
counts, where each entry $y_{ij}$ corresponds to the number of individuals of
species $j \in \{1,\dots,m\}$ in sequence $i \in \{1,\dots,n\}$. Similarly, let
$\widehat{\mathbf{Y}} \in \{0,1,2,\dots\}^{n \times m}$ denote the matrix of
predicted counts. The MCRMSE is defined as:
\begin{equation}\label{eq:mcrmse}
 \mathit{MCRMSE} = \frac{1}{m} \sum\limits_{j=1}^{m} \sqrt{\frac{1}{n}
\sum\limits_{i=1}^{n} (y_{ij} - \hat y_{ij})^2}.
\end{equation}
This metric accounts for both the species identification and counting
errors, ensuring that false predictions on empty sequences also contribute to
the overall error.

\autoref{tab:iwildcam2021_results} presents the final standings based on the
private score, which was calculated using 50\% of the test set. Our approach was
ranked first in the challenge out of 42 teams that entered the competition.

\begin{table}[t]
    \centering
    \caption{Final results on the iWildCam 2021 private test set. Lower scores
are better.}
    \label{tab:iwildcam2021_results}
    \begin{tabular}{l c}
        \toprule
        \textbf{Method} & \textbf{Private Score} \\
        \midrule
        \textbf{1st place (Ours)}                & \textbf{0.0293278} \\
        2nd place                               & 0.0293786 \\
        3rd place                               & 0.0308570 \\
        4th place                               & 0.0308756 \\
        5th place                               & 0.0336513 \\
        6th place                               & 0.0339028 \\
        iWildCam 2020 winning (baseline)        & 0.0348171 \\
        7th place                               & 0.0376347 \\
        8th place                               & 0.0383956 \\
        All-zero baseline                       & 0.0385147 \\
        \bottomrule
    \end{tabular}
\end{table}

\begin{table*}[t]
    \centering
    \caption{Ablation study of the proposed iWildCam 2021 solution. Lower
scores are better. The final configuration is
    highlighted in bold.}
    \label{tab:iwildcam2021_baseline_ablation}
    \begin{tabular}{l l c c}
        \toprule
        \textbf{Test-Time Aug.} & \textbf{Seq. Aggreg.} &
        \textbf{MDv4 Threshold} & \textbf{Private Score} \\
        \midrule
        None              & Voting  & 0.9 & 0.0314948 \\
        Horizontal flip   & Voting  & 0.9 & 0.0305960 \\
        Horizontal flip   & Average & 0.9 & 0.0294853 \\
        \textbf{Horizontal flip} &
        \textbf{Average} &
        \textbf{0.8} &
        \textbf{0.0293278} \\
        \bottomrule
    \end{tabular}
\end{table*}

As stated by the competition organizers \citep{beery2021iwildcam}, the MCRMSE
metric tends to be a small number even when the error in counts is large. This
occurs because camera traps typically have a small number of individuals and
because the model is double penalized for both species and counting errors. The
organizers provided some simple baselines, which we specify in
\autoref{tab:iwildcam2021_results} for comparison purposes. One baseline
uses the iWildCam 2020 winner's species predictions combined with the maximum
number of bounding boxes with confidence score higher than 0.8 across the
sequence -- in line with our solution. The all zeros baseline predicts zero for
all instances, which performs surprisely well. This highlights how challenging
this counting problem is, given its inherent conditions.

We present an ablation study to demonstrate the effectiveness of certain design
choices in our heuristic solution, as shown in
\autoref{tab:iwildcam2021_baseline_ablation}. This study considers only the
species classifier at the image level used for the final submission. A
well-known approach to improving classification predictions is using test-time
augmentation; in our solution, we added a horizontal flip to both full image and
bbox models. We also found that averaging the species predictions across images
in a sequence provides better performance than the majority voting approach.
Finally, we used a threshold of $0.8$ for counting MegaDetectorV4 bounding
boxes, as proposed by the organizers. We initially used $0.9$ during the
competition but found that it discarded too many valid bounding boxes.

%% file: sec/4_conclusion.tex
\section{Conclusion}

Counting animals in sequences of camera trap images is a challenging problem,
especially considering the scenario where there are no count labels, as is the
case for most datasets. In this work, we presented a solution based on a
heuristic strategy that uses the bounding boxes from MegaDetectorV4 combined
with an ensemble of classifiers. Although our strategy is limited to predicting
only one species per sequence, it has been proven to be a strong baseline for
the problem, winning the iWildCam 2021 challenge.

However, we believe that a more natural mechanism should be based on tracking
animals across images (multi-object tracking), classifying each track, and
counting them. Such an approach could also be more interpretable to humans,
avoiding ``black box'' solutions. During the competition, we experimented
with DeepSORT \citep{Wojke2017simple} and variations using embedding features
from ReID models. However, due to time constraints, we were unable to explore
this strategy in depth.

\section*{Acknowledgments}
This study was financed in part by the Coordenação de Aperfeiçoamento de Pessoal
de Nível Superior - Brasil (CAPES) - Finance Code 001. This work was partially
supported by Amazonas State Research Support Foundation - FAPEAM -  through the
POSGRAD project.